 \documentclass[pmlr,twocolumn]{jmlr} 

\PassOptionsToPackage{square, numbers}{natbib}

\usepackage{booktabs}
\usepackage[load-configurations=version-1]{siunitx} 

\theorembodyfont{\upshape}
\theoremheaderfont{\scshape}
\theorempostheader{:}
\theoremsep{\newline}

\jmlrvolume{ML4H Extended Abstract arXiv Index}
\jmlryear{2020}
\jmlrsubmitted{2020}
\jmlrpublished{}
\jmlrworkshop{Machine Learning for Health (ML4H) 2020} 

\title[Event-Based Hidden Markov Modelling]{Learning transition times in event sequences: the Event-Based Hidden Markov Model of disease progression}

 \author{%
  \Name{Peter A. Wijeratne}\textnormal{*}\and
  \Name{Daniel C. Alexander} \\
  \addr Centre for Medical Image Computing, Department of Computer Science, University College London, Malet Place, London WC1E 6BT\\
  \Email{*p.wijeratne@ucl.ac.uk} 
 }
\begin{document}
\maketitle
\begin{abstract}
Progressive diseases worsen over time and are characterised by monotonic change in features that track disease progression. Here we connect ideas from two formerly separate methodologies -- event-based and hidden Markov modelling -- to derive a new generative model of disease progression. Our model can uniquely infer the most likely group-level sequence and timing of events (\textit{natural history}) from limited datasets. Moreover, it can infer and predict individual-level trajectories (\textit{prognosis}) even when data are missing, giving it high clinical utility. Here we derive the model and provide an inference scheme based on the expectation maximisation algorithm. We use clinical, imaging and biofluid data from the Alzheimer's Disease Neuroimaging Initiative to demonstrate the validity and utility of our model. First, we train our model to uncover a new group-level sequence of feature changes in Alzheimer's disease over a period of ${\sim}17.3$ years. Next, we demonstrate that our model provides improved utility over a continuous time hidden Markov model by area under the receiver operator characteristic curve ${\sim}0.23$. Finally, we demonstrate that our model maintains predictive accuracy with up to $50\%$ missing data. These results support the clinical validity of our model and its broader utility in resource-limited medical applications.
\end{abstract}
\section{Introduction}
\label{sec:intro}
Progressive diseases such as Alzheimer's disease (AD) are characterised by monotonic deterioration in functional, cognitive and physical abilities over a period of years to decades \cite{Masters2015}. AD has a long prodromal phase before symptoms become manifest (${\sim}20$ years), which presents an opportunity for therapeutic intervention if individuals can be identified at an early stage in their disease trajectory \cite{Dubois2016}. Clinical trials for disease-modifying therapies in AD would also benefit from methods that can stratify participants, both in terms of individual-level disease stage and rate of progression \cite{Cummings2019}.

Data-driven models of disease progression can be used to learn hidden information, such as individual-level stage, from observed data \cite{Oxtoby2017}. In this paper we address the problem of how to learn transition times in event sequences of disease progression, which is a long-standing problem in the methods community \cite{Huang2012,Fonteijn2012}. The solution to this problem also has clinical demand, as it provides the basis for an interpretable timeline of disease progression events that can be used for prognosis. We connect ideas from two formerly separate methodologies -- event-based and hidden Markov modelling -- to derive a new generative event-based hidden Markov model (EB-HMM) of disease progression. As such, this paper has three main novelties:
\begin{enumerate}
\item it generalises a formerly cross-sectional model (the EBM: event-based model \cite{Fonteijn2012}), allowing it to accommodate longitudinal data;
\item it defines a Bayesian `event-based' framework to inject prior information into structured inference from longitudinal data, allowing us to learn from limited datasets;
\item it uses EB-HMM to learn a new clinically interpretable sequence and timing of events in Alzheimer's disease (\textit{natural history}), and to predict individual-level trajectories (\textit{prognoses}).
\end{enumerate}
EB-HMM has strong clinical utility, as it provides an interpretable group-level model of how features of disease progression (\textit{biomarkers}) change over time. Such a model for AD was first hypothesised by \cite{Jack2013}, but EB-HMM is the first to provide a single, unified methodology for learning data-driven sequences and timing of events in progressive diseases. Moreover, EB-HMM naturally handles missing data; both in terms of partially missing data (when an individual does not have measurements for every feature) and completely missing data (when an individual is not observed at a given time-point). This capability gives EB-HMM broad utility in clinical practice, particularly in resource-limited scenarios (e.g., small hospitals) where medical practitioners may not have access to a complete set of measurements. Finally, EB-HMM also advances on AI-driven clinical trial design, where model-derived information could be used to inform biomarker and cohort selection criteria \cite{Dorsey2015}.
\section{Methods}
\label{sec:theory}
\subsection{Event-Based Hidden Markov Model}
To formulate EB-HMM, we make three assumptions, namely $i)$ monotonic biomarker change; $ii)$ a consistent event sequence, $S$, across the whole sample; and $iii)$ Markov (memoryless) stage transitions. The model likelihood is:
\begin{align}
\begin{split}
P(Y \vert \theta, S) = \prod_{j=1}^{J} \left[ \sum_{k=0}^{N} P(k_{j, t=0}) \prod_{t=1}^{T_j} P(k_{j,t} \vert k_{j,t-1}) \right. \\
    \left. \prod_{t=0}^{T_j} \prod_{i=1}^{k_{j,t}} P(Y_{i, j,t} \vert k_{j,t}, \theta^p_i, S) \right. \\ \left. \prod_{i=k_{j,t}+1}^I P(Y_{i, j,t} \vert k_{j,t}, \theta^c_i, S) \right].
    \label{eq:tebm_like}
\end{split}
\end{align}
For a full derivation of Equation \ref{eq:tebm_like} and descriptions of each variable see Appendix \ref{sec:app:theory}. We then make the usual Markov assumptions to obtain the form of the $N \times N$ dimensional transition generator matrix $Q_{a, b}$:
\begin{align}
\begin{split}
    exp(\Delta Q)_{a,b} = P(k_{j,t}=a \vert k_{j,t-1 } = b, \Delta).
\end{split}
\end{align}
Here we have assumed a homogeneous continuous-time process $\tau$, and that the state duration $\Delta=\tau_t - \tau_{t-1}$ is (matrix) exponentially distributed, $\Delta \sim exp(\Delta)$, between states $a,b$. The former follows from our original assumption that the sequence $S$ (and hence $Q$) is independent of time, and the latter is a solution to the rate equation. The $N$ dimensional initial state probability vector $\pi_a$ is defined as:
\begin{align}
\begin{split}
    \pi_a = P(k_{j,t=0}=a).
\end{split}
\end{align}
Finally, the expected duration of each state (sojourn time), $\Delta_k$, is given by~\cite{Rabiner1989}:
\begin{align}
\begin{split}
    \Delta_k = \sum_{\Delta=1}^{\infty} \Delta p_k(\Delta) = \frac{1}{1-q_{kk}}.
    \label{eq:tebm_duration}
\end{split}
\end{align}
Here $p_k(\Delta)$ is the probability density function of $\Delta$ in state $k$, and $q_{kk}$ are the diagonal elements of the transition matrix $Q_{a,b}$.
\subsection{Inference}
We aim to learn the sequence $\overline{S}$, initial probability $\overline{\pi}_a$, and transition matrix $\overline{Q}_{a,b}$, that maximise the complete data log likelihood, $\mathcal{L(\overline{S},\overline{\pi},\overline{Q})} = \mathrm{log} P(Y \vert S, \pi, Q; \theta)$. The overall inference scheme is shown in Appendix \ref{sec:app:inf}.
\subsection{Staging}
After fitting $\overline{S}$, $\overline{\pi}_a$ and $\overline{Q}_{a,b}$, we infer the most likely Markov chain (i.e., trajectory) for each individual using the standard Viterbi algorithm \cite{Rabiner1989}. We can also use EB-HMM to predict individual-level future stage by multiplying the transition matrix, $\overline{Q}_{a,b}$, with the posterior probability for the individual at time $t$, and selecting the maximum likelihood stage:
\begin{align}
\begin{split}
    \mathrm{arg\,max}_k P(Y_{t+1} | k_b; \overline{S}) = \\ \mathrm{arg\,max}_k P(Y_{t} | k_a; \overline{S}) \cdot \overline{Q}_{a,b}.
    \label{eq:tebm_pred}
\end{split}
\end{align}
\section{Results}
\label{sec:res}
\subsection{Alzheimer's disease timeline}
\label{sec:time}
We use EB-HMM to infer the group-level sequence of events and the time between them in the ADNI cohort. Figure \ref{fig:timeline} shows the corresponding order and timeline of events, and baseline and predicted stages estimated by EB-HMM for two representative patients. For descriptions of the data and the model training scheme see Appendix \ref{sec:app:data} and \ref{sec:app:train}. This timeline is the first of its type in the field of AD progression modelling, and reveals a chain of observable events occurring over a period of ${\sim}17.3$ years. The ordering largely agrees with previous model-based analyses \cite{Young2014,Oxtoby2018}, and EB-HMM provides additional information on the time between events. Early changes in biofluid measures (ABETA, TAU, PTAU) over a short timescale have been proposed in a recent hypothetical model of AD biomarker trajectories \cite{Jack2013}. Early observable change in the brain (represented here by the ventricles) is also reported, followed by a chain of cognitive and structural changes, with change in the whole brain volume occurring last.
\begin{figure*}[htbp]
\floatconts
  {fig:timeline}
  {\caption{AD timeline inferred by EB-HMM. The order of events on the horizontal axis is given by $\overline{S}$, and the time between events is calculated from $\overline{Q}$. Baseline stage (solid arrow) and predicted next stage (shaded arrow) estimated by EB-HMM for two example patients are shown, chosen from the MCI and AD sub-groups.}}
  {\includegraphics[width=\linewidth]{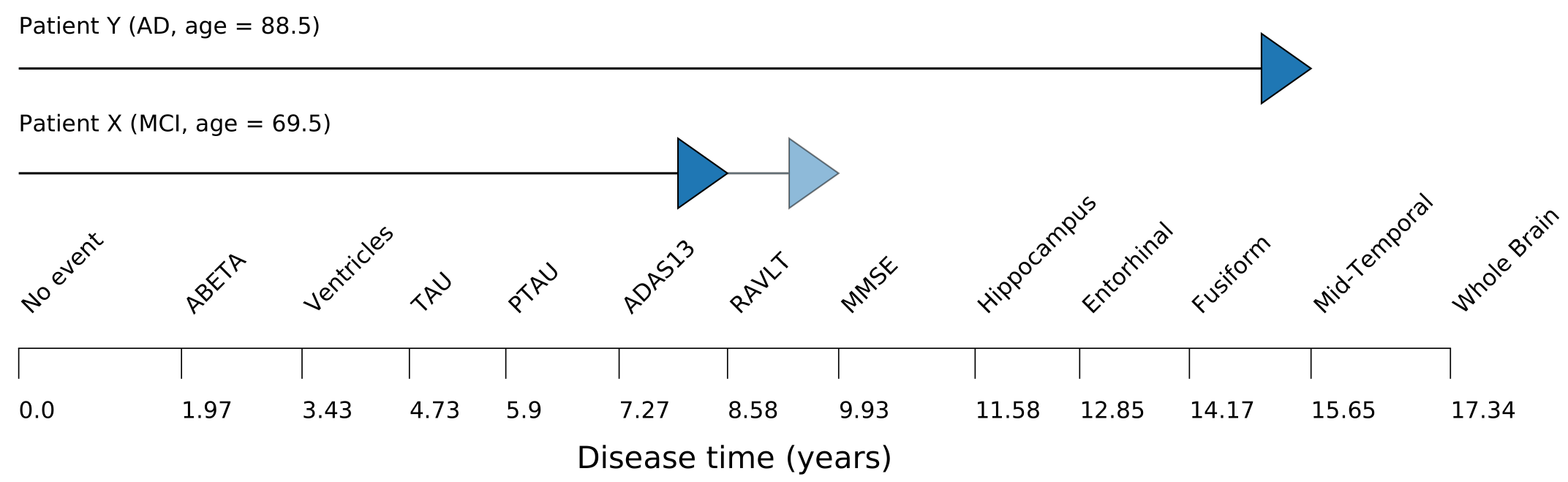}}
\end{figure*}
\subsection{Comparative model performance}
\label{sec:comp}
We train EB-HMM and a continuous-time hidden Markov model (CT-HMM) to infer individual-level stage sequences and hence compare predictive accuracy on a common task. Specifically, we use baseline stage as a predictor of conversion from CN to MCI, or MCI to AD, over a period of two years. Here predicted converters are defined as people with a stage greater than a threshold stage, which is iterated across all possible stages. We calculate the area under receiver operating characteristic curve (AU-ROC), and perform 5-fold cross-validation. Table \ref{tab:comp} shows that EB-HMM performs substantially better than CT-HMM in both the full (including individuals with missing data) and subset data (only individuals with complete data).
\vspace{-.5cm}
\begin{table}[htbp]
\floatconts
  {tab:comp}
  {\caption{EB-HMM and CT-HMM performance for the task of predicting conversion, using either the full or subset data.\vspace{-.5cm}}}%
  {%
    \begin{tabular}{|l|l|l|}
    \hline
    \abovestrut{2.2ex}\bfseries Model & \bfseries AU-ROC\\\hline
    EB-HMM (full) & $\mathbf{0.804\pm0.07}$\\
    EB-HMM (subset) & $0.737\pm0.09$\\
    CT-HMM (subset) & $0.579\pm0.12$\\\hline
    \end{tabular}
  }
\end{table}
\vspace{-.5cm}
\subsection{Performance with missing data}
\label{sec:miss}
Finally, we demonstrate the ability of EB-HMM to handle missing data. We randomly discard $25\%$, $50\%$, and $75\%$ of the feature data from each individual in the subset data and re-train EB-HMM. As in \sectionref{sec:comp}, we use prediction of conversion as the task and 5-fold cross-validation to obtain out-of-sample estimates of the AU-ROC. Table \ref{tab:miss} shows that EB-HMM maintains consistent performance up to $50\%$ missing data, and drops off only moderately for $75\%$.
\vspace{-.5cm}
\begin{table}[htbp]
\floatconts
  {tab:miss}
  {\caption{EB-HMM performance for the task of predicting conversion with missing data.\vspace{-.5cm}}}%
  {%
    \begin{tabular}{|l|l|}
    \hline
    \abovestrut{2.2ex}\bfseries $\%$ missing & \bfseries AU-ROC\\\hline
    $25\%$ & $0.722\pm0.09$\\
    $50\%$ & $0.719\pm0.13$\\
    $75\%$ & $0.669\pm0.15$\\\hline
    \end{tabular}
  }
\end{table}
\vspace{-.5cm}
\section{Discussion}
\label{sec:diss}
Future work on EB-HMM will be focused on relaxing its assumptions\footnote{While the requirement of a control sample for fitting the EB-HMM mixture model distributions could be deemed a limitation, it is arguably a strength as it allows us to informatively leverage control data; a key issue that was highlighted by \cite{Wang2014}.}, namely $i)$ monotonic biomarker change; $ii)$ a consistent event sequence across the whole sample; and $iii)$ Markov (memoryless) stage transitions. Assumption $i)$ is both a limitation and a strength: it allows us to simplify our model at the expense of requiring monotonic biomarker change; as shown here, for truly monotonic clinical, imaging and biofluid markers it only provides benefits. However for non-monotonic markers -- such as heart rate -- either the model or data would need to be adapted. Assumptions $ii)$ and $iii)$ could be relaxed by combining our EB-HMM framework with (for example) subtype modelling \cite{Young2018} and semi-Markov modelling \cite{Alaa2018}, respectively. In particular, EB-HMM can be directly integrated into the subtyping and staging framework proposed by \cite{Young2018}, which would allow us to capture the well-reported heterogeneity in AD and produce timelines such as Figure \ref{fig:timeline} for separate subtypes. This opens up the prospect of developing a probabilistic model that can infer interpretable longitudinal subtypes from limited datasets.
\vspace{-.5cm}
\acks{We thank all the ADNI study participants.}
\bibliography{jmlr-sample}
\appendix
\section{Appendix}
\label{sec:app}
\subsection{Event-Based Hidden Markov Model}
\label{sec:app:theory}
We can write the EB-HMM joint distribution over all variables in a hierarchical Bayesian framework:
\begin{align}
\begin{split}
    P(S, \theta, k, Y) = P(S) \cdot P(\theta \vert S) \\ \cdot P(k \vert \theta, S) \cdot P(Y \vert k, \theta, S).
    \label{eq:tzsm_joint}
\end{split}
\end{align}
Here $S$ is the hidden sequence of events, $\theta$ are the distribution parameters generating the data, $k$ is the hidden disease state, and $Y$ are the observed data. Graphical models of CT-HMM and EB-HMM are shown in Figure~\ref{fig:graphs}. Note that we have assumed conditional independence of $S$ from $k$; that is, the complete set of disease progression states is independent of the time of observation.
\begin{figure}[htbp]
\floatconts
  {fig:graphs}
  {\caption{Graphical models for (a) CT-HMM and (b) EB-HMM. Hidden variables are denoted by circles, observations by squares. $S$: sequence of events; $\theta$: distribution parameters; $k$: disease state; $Y$: observed data; $T$: observed time.}}
  {%
    \subfigure[CT-HMM]{\label{fig:circle}%
      \includegraphics[width=0.4\linewidth]{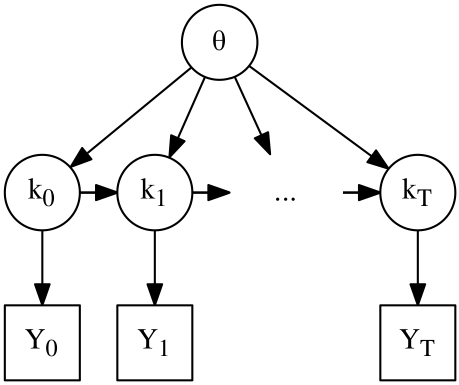}}%
    \qquad
    \subfigure[EB-HMM]{\label{fig:square}%
      \includegraphics[width=0.4\linewidth]{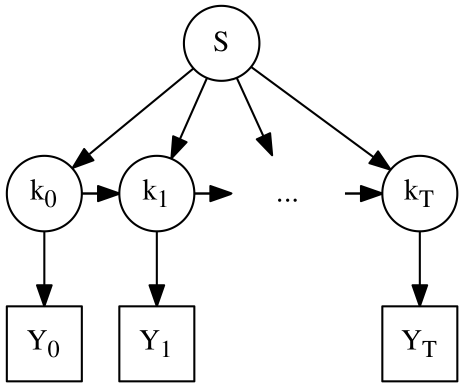}}
  }
\end{figure}
\vspace{-0.5cm}
Assuming independence between observed features $i=1,...,I$, if a patient $j=1,...,J$ is at latent state $k_{j,t} = 0,...,N$ at time $t=1,...,T_j$ in the progression model, the likelihood of their data $Y_{j,t}$ is given by:
\begin{align}
\begin{split}
    P(Y_{j,t} \vert k_{j,t}, \theta, S) =  \prod_{i=1}^{I} P(Y_{i, j,t} \vert k_{j,t}, \theta_i, S).
    \label{eq:tzsm_like_individual}
\end{split}
\end{align}
Here $\theta_i$ are the distribution parameters for feature $i$, defined by a hidden sequence of events $S = (s(1),...,S(N))$. Following \cite{Fonteijn2012}, we enforce the monotonicity hypothesis by requiring $S$ to be ordered, meaning individuals at stage $k_{j,t}$ cannot revert back to an earlier stage. This assumption is necessary to allow snapshots from different individuals to inform on the full event ordering. Next, we assume a Markov jump process between discrete time-points:
\begin{align}
\begin{split}
    P(Y_{j} \vert k_{j}, \theta, S) = P(k_{j, t=0}) \prod_{t=1}^{T_j} P(k_{j,t} \vert k_{j,t-1}) \\
    \prod_{t=0}^{T_j} \prod_{i=1}^{I} P(Y_{i, j,t} \vert k_{j,t}, \theta_i, S).
    \label{eq:tzsm_like_markov}
\end{split}
\end{align}
To obtain an event-based model, we now define prior values for the distribution parameters $\theta$ for each state $k$ in sequence $S$. Following \cite{Fonteijn2012} we choose a two-component Gaussian mixture model to describe the data likelihood:
\begin{align}
\begin{split}
\label{eq:gmm}
   \prod_{i=1}^{I} P(Y_{i, j,t} \vert k_{j,t}, \theta_i, S) = \prod_{i=1}^{k_{j,t}} P(Y_{i, j,t} \vert k_{j,t}, \theta^p_i, S) \\ \prod_{i=k_{j,t}+1}^I P(Y_{i, j,t} \vert k_{j,t}, \theta^c_i, S)
\end{split}
\end{align}
Here $\theta^p_i=[\mu^p_i,\sigma^p_i,w^p_i]$ and $\theta^c_i=[\mu^c_i,\sigma^c_i,w^c_i]$ are the mean, $\mu$, standard deviation, $\sigma$, and mixture weights, $w$, for the patient and control distributions, respectively. Note that these distributions are fit prior to inference, which requires our data to contain labels for patients and controls; however, once $\theta^p_i$ and $\theta^c_i$ have been fit, the model can infer $S$ without any labels. One of the strengths of the mixture model approach is that when feature data are missing, the two probabilities on the RHS of Equation \ref{eq:gmm} can simply be set equal.

To obtain the total data likelihood, we marginalize over the hidden state $k$ and assume independence between measurements from different individuals $j$ (dropping indices $j,t$ in the sum for notational simplicity):
\begin{align}
\begin{split}
P(Y \vert \theta, S) = \prod_{j=1}^{J} \left[ \sum_{k=0}^{N} P(k_{j, t=0}) \prod_{t=1}^{T_j} P(k_{j,t} \vert k_{j,t-1}) \right. \\
    \left. \prod_{t=0}^{T_j} \prod_{i=1}^{k_{j,t}} P(Y_{i, j,t} \vert k_{j,t}, \theta^p_i, S) \right. \\ \left. \prod_{i=k_{j,t}+1}^I P(Y_{i, j,t} \vert k_{j,t}, \theta^c_i, S) \right].
    \label{eq:tzsm_like_final}
\end{split}
\end{align}
We can now use Bayes' theorem to obtain the posterior distribution over $S$. We note that Equation~\ref{eq:tzsm_like_final} is the time generalisation of the model presented by~\citep{Fonteijn2012}, and for $T_j=1$ it reduces to that model.  We further note that Equation~\ref{eq:tzsm_like_markov} looks like a CT-HMM ~\cite{Ghahramani2001}. The mathematical innovation of our work is to reformulate the EBM in a CT-HMM framework\footnote{Or conversely, the CT-HMM in an event-based framework.}. To our knowledge this is the first such model of its type.
\subsection{Inference scheme}
\label{sec:app:inf}
We use a nested inference scheme based on iteratively optimising the sequence $S$, and fitting the initial probability $\pi_a$ and transition matrix $Q_{a,b}$, to find a local maximum via a nested application of the Expectation-Maximisation (EM) algorithm. At the first EM step, $S$ is optimised for the current values of the initial probability $\pi'_a$ and transition matrix $Q'_{a,b}$, by permuting the position of every event separately while keeping the others fixed. At the second step, $\pi_a$ and $Q_{a,b}$ are fitted for the current sequence $S'$ using the standard forward-backward algorithm~\cite{Rabiner1989}. Here we apply only a single pass, as iterative updating of $\pi_a$ and $Q_{a,b}$ while keeping $S$ (and hence $\theta_i$) fixed effectively turns the optimisation problem into repeated scaling of the posterior, which causes over-fitting of $\pi_a$ and $Q_{a,b}$.
\begin{algorithm2e}
\caption{EB-HMM inference}
\label{alg:algo}
\SetKwInOut{Input}{Input}\SetKwInOut{Output}{Output}
\Input{$Y$}
\Output{$S$, $\pi$, $Q$}
Initialise $S$\;
\While{not $S$ converged}{
    \tcp{E-step of sequence optimisation}
    \While{not every event permuted}{
        Initialise $\pi$, $Q$\;
        \tcp{E-step of transition and initial probability optimisation}
        Compute $\gamma_{a,t} = P(k_{t} = a \vert Y, S=S'; \pi, Q)$\;
        Compute $\xi_{a,b,t} = P(k_t = a, k_{t+1} = b \vert Y, S=S'; Q)$\;
        \tcp{M-step of transition and initial probability optimisation}
        Update $\pi_a \leftarrow \gamma_{a,0}$\;
        Update $Q_{a,b} \leftarrow \dfrac{\sum_{t=1} \xi_{a,b,t}}{\sum_{t=1} \gamma_{a,t-1}}$\;
        Compute $\mathcal{L}(S) = \mathrm{log} P(Y \vert S; \pi', Q')$\;
     }
    \tcp{M-step of sequence optimisation}
    Update $S \leftarrow \mathrm{arg\,max}_S\mathcal{L}(S)$\;
 }
\end{algorithm2e}
\vspace{-0.5cm}
\subsection{Alzheimer's disease data}
\label{sec:app:data}
We use data from the ADNI study, a longitudinal multi-centre observational study of AD \cite{Mueller2005}. We select 468 participants (119 CN: cognitively normal; 297 MCI: mild cognitive impairment; 29 AD: manifest AD; 23 NA: not available), and three time-points per participant (baseline and follow-ups at 12 and 24 months). Individuals were allowed to have missing data at any time-point. Note that we use a subset of 368 individuals with no missing data in Sections \ref{sec:comp} and \ref{sec:miss}. We train on a mix of 12 clinical, imaging and biofluid features. The clinical data are three cognitive markers: ADAS-13,
Rey Auditory Verbal Learning Test (RAVLT) and Mini-Mental State
Examination (MMSE). The imaging data are T1-weighted 3T structural magnetic resonance imaging (MRI) scans, post-processed to produce regional volumes using the GIF software tool \cite{Cardoso2015}. We select a subset of sub-cortical and cortical regional volumes with reported sensitivity to AD pathology, namely the hippocampus, ventricles, entorhinal, mid-temporal, and fusiform, and the whole brain \cite{Frisoni2010}. The biofluid data are three cerebrospinal fluid markers: amyloid-$\beta_{1-42}$ (ABETA), phosphorylated tau (PTAU) and total tau (TAU). The TADPOLE challenge dataset \cite{Marinescu2020} used in this paper is freely available upon registering with an ADNI account.
\vspace{-0.1cm}
\subsection{Model training}
\label{sec:app:train}
We compare EB-HMM and CT-HMM algorithms. To ensure fair comparison, we impose a constraint on both models by placing a 2nd order forward-backward prior on the transition matrix. For EB-HMM, we fit Gaussian mixture models to the distributions of AD (patients) and CN (controls) sub-groups prior to running Algorithm \ref{alg:algo}. For CT-HMM, we apply the standard forward-backward algorithm and iterate the likelihood to convergence within $10^{-2}$ of the total model likelihood. We initialise the CT-HMM prior mean and covariance matrices from the training data, using standard $k$-means and the feature covariance, respectively. EB-HMM is implemented and parallelised in Python; open-source code will be provided upon full journal publication at the author's repository: \url{https://github.com/pawij/tebm}.
\end{document}